\documentclass[11pt,a4paper]{article}
\usepackage[hyperref]{eacl2021}
\usepackage{times}
\usepackage{latexsym}

\usepackage{microtype}

\aclfinalcopy % Uncomment this line for the final submission
\usepackage{graphicx}
\usepackage{amsmath}
\usepackage{multirow}
\usepackage{subcaption}
\usepackage{xcolor}

\definecolor{slot-blue}{RGB}{0, 86, 240}
\definecolor{slot-green}{RGB}{24, 130, 0}
\definecolor{slot-orange}{RGB}{255, 100, 0}

\title{Attention Guided Dialogue State Tracking with Sparse Supervision}

\author{Shuailong Liang \textsuperscript{1}, Lahari Poddar\textsuperscript{2}, Gyuri Szarvas\textsuperscript{2}\\
\textsuperscript{1}Singapore University of Technology and Design\\, \textsuperscript{2}Amazon Development Center Germany GmbH\\
 liangshuailong@gmail.com \{poddarl,szarvasg\}@amazon.com}

\date{}

\begin{document}
\maketitle

\begin{abstract}
Existing approaches to Dialogue State Tracking (DST) rely on turn level dialogue state annotations, which are expensive to acquire in large scale. 
In call centers, for tasks like managing bookings or subscriptions, the user goal can be associated with actions (e.g.~API calls) issued by customer service agents. % during the conversation. 
These action logs are available in large volumes and can be utilized for learning dialogue states. 
However, unlike turn-level annotations, such logged actions are only available sparsely across the dialogue, providing only a form of weak supervision for DST models.

To efficiently learn DST with sparse labels, 
we extend a state-of-the-art encoder-decoder model. 
The model learns a slot-aware representation of dialogue history, which focuses on relevant turns to guide the decoder.
 We present results on two public multi-domain DST datasets (MultiWOZ and Schema Guided Dialogue) in both settings i.e. training with  {\it turn-level} and with {\it sparse} supervision.
The proposed approach improves over baseline in both settings. 
 More importantly, our model  trained with sparse supervision is competitive in performance to fully supervised baselines, while being more data and cost efficient.% and converging much faster.

\end{abstract}
\section{Introduction}
\label{intro}

%Conversational AI systems have recently seen a surge of research interest not only in academia but also in industry. 
A task-oriented conversational system helps a user achieve a specific goal, such as booking a hotel, taxi, or a table at a restaurant \cite{gao2018neural}. 
An essential component in dialogue systems is dialogue state tracking (DST), i.e. the accurate prediction of a user's goal as the dialogue progresses. 
It is typically modelled as slot (\textit{time})-value (\textit{18:30}) pairs specific to a domain (\textit{taxi}). 
Current DST systems rely on extensive human annotation of slot-values at each dialogue turn as supervision signal for training. 
This is prone to annotation errors (c.f. the extensive label corrections between MultiWOZ 2.0 and 2.1 \cite{eric2019multiwoz}) and costs a lot of time and effort to create.
Furthermore, domains and services emerge continuously.
Generalization to new domains using limited data is one of the most important challenges for using DST and goal-oriented conversational AI research more broadly.

\begin{figure*}[t]
		\centering
		\begin{subfigure}{0.95\linewidth}
	\includegraphics[width=\linewidth]{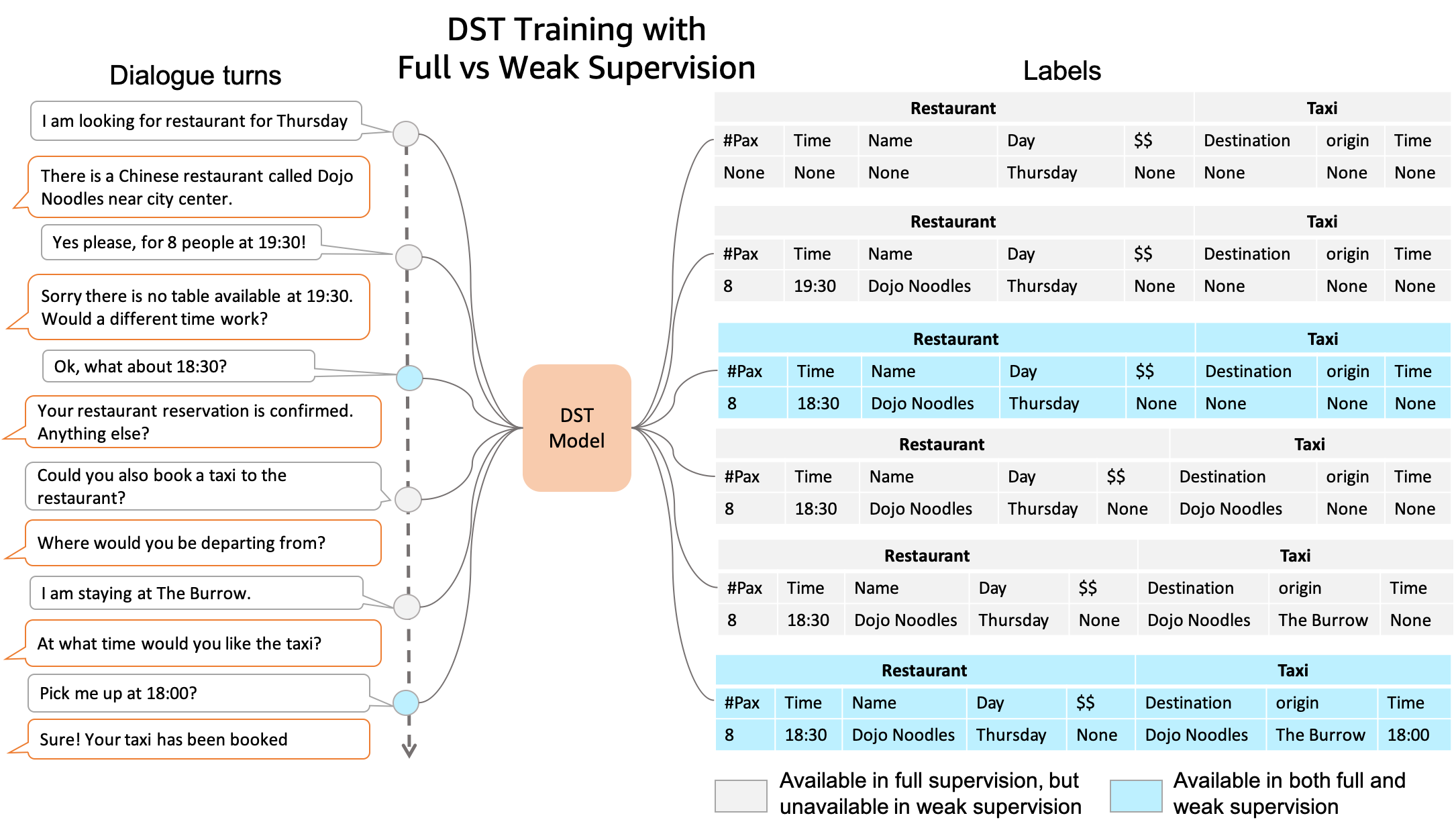}
	\caption{}
	\label{fig:training-dst}
	\end{subfigure}
\vspace{0.2cm}

		\begin{subfigure}{0.9\linewidth}
	\includegraphics[width=\linewidth]{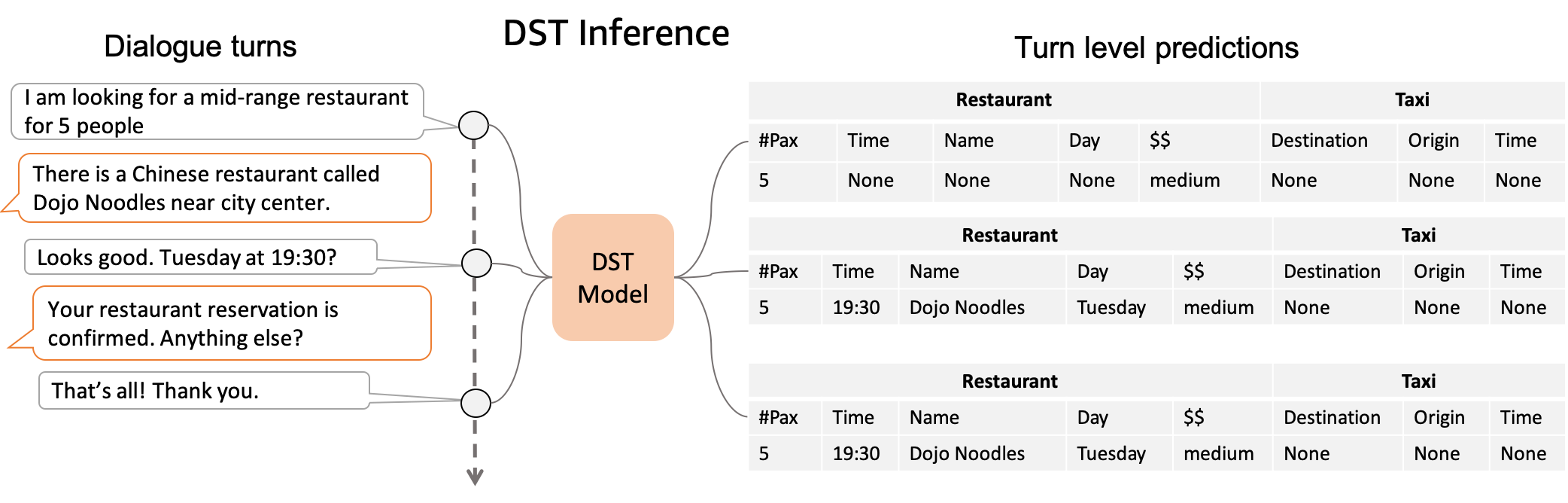}
		\caption{}
			\label{fig:pred-dst}
\end{subfigure}
\caption{
	Illustration of  DST training and inference procedure. (a) depicts the difference between training a model with full vs weak supervision signals. In case of weak supervision, only the labels for a few intermediate turns (in blue) are available. For full supervision, labels for all turns are available.  (b) shows the inference setting for DST. The model needs to make turn level predictions irrespective of how it was trained.}
	\label{fig:api_calls_as_weak_supervision}
	\vspace{-0.3cm}
\end{figure*}

We propose a novel approach of learning DST through sparse supervision signals, which can be obtained in large quantities without manual annotation. 
In call-centers, customer service agents handle a large number of user queries every day. 
%These  human-to-human dialogues can be utilized to learn dialogue response models.
The actions taken by agents while resolving user queries are reflected in logged API function calls, together with associated parameters and their values. 
An example API call to refund a customer for an order could be: 
{\tt issue\_refund(order\_id, refund\_method, amount)}, where {\tt order\_id} is the unique ID of the order, {\tt refund\_method} could be credit card or voucher, and {\tt amount} is the money to be refunded.
This is analogous to the notion of {\it domain} and associated {\it slots} and {\it values} in public DST datasets and represent the state of a dialogue (the goal of dialogues being to issue those API calls). 
%Such logged external API calls are available at practically no additional cost and can be obtained in large volumes. We argue that these logs can lend themselves as a natural proxy to the dialogue state (the goal of dialogues being to issue those API calls). 
However, unlike annotated dialogue states, these calls would only be available sparsely across the turns, making it challenging to learn a DST model. 

Figure~\ref{fig:api_calls_as_weak_supervision} illustrates the difference between full supervision with manual annotation, and weak supervision acquired from sparsely logged API calls. 
As can be seen from Figure~\ref{fig:training-dst}, training with full supervision provides slot labels at every turn, while in weakly supervised setting the model would have labels only for a few turns where a request is confirmed and an API call (e.g. {\tt book\_restaurant}) is issued.
Irrespective of how a model is trained, 
during inference the dialogue state would need to be predicted at every turn (Figure \ref{fig:pred-dst}) ,
%to build an end-to-end conversational system, learning a turn-level DST model is still desirable, such that the predicted dialogue state can be used 
to inform the response generation and resolve the task at the appropriate turn. 
In other words, with weak supervision, labels are provided sparsely during training, while dialogue state predictions must be made for every turn in a dialogue.
%Therefore, the evaluation is still at the turn level for the weak-supervision setting.

Our main contributions are the following. 

(1) We define a novel learning setup, where labels for only a few dialogue turns are used to learn a DST model with weak supervision. 
Despite the challenging nature of the task, 
this is a practical setting 
%compared to domain adaptation or zero-shot learning which have been studied in recent literature, 
that leverages large scale, robust, and cheap training data available for a new domain.

(2) We propose a neural encoder-decoder architecture, named AGG, with an attention mechanism to help the model focus on small and important fractions of the input. This enables the model to achieve superior performance in the weakly supervised setting. 

Our experiments on the public benchmark MultiWOZ 2.1 and Schema Guided Dialogues datasets show that under weak supervision, our proposed model achieves a joint goal accuracy improvement of 5.94\% (MultiWOZ 2.1), and 4.95\% (Schema Guided Dialogues) absolute over the baseline model. Further analysis shows that on MultiWOZ 2.1 dataset, using sparse supervision is comparable to training the baseline model using full supervision on 60\% of the dialogues, which proves that our proposed model is particularly data efficient. 

To the best of our knowledge, this is the first work to study DST in a weakly supervised setting, with sparse labels for dialogue states, that aligns well with practical applications of Conversational AI. 

\section{Related Work}

Multidomain DST has gained recent attention with the availability of large scale datasets such as MultiWOZ 2.0 ~\cite{budzianowski-etal-2018-multiwoz} and its successor MultiWOZ 2.1 ~\cite{eric2019multiwoz}, which focus on restaurant, train, taxi, attraction and hotel domains, and Schema Guided Dialogues Dataset ~\cite{rastogi2019towards}.

The recent works on DST can be categorized into 1) Picklist; 2) Span prediction; and 3) Generation based models. Picklist models \cite{zhang2019find} depend heavily on an ontology, which is hard to keep updated, especially for slots that have a large and dynamic range of values (e.g. restaurant names). Span prediction based approaches \cite{rastogi2019towards} are used extensively in machine reading comprehension literature ~\cite{rajpurkar-etal-2016-squad} but some slot values can not be directly copied from source side, such as ``yes or no" questions. 
Additionally, it becomes even harder to find an exact match if the dataset has been created by annotators writing down the slot value instead of marking the spans. 
To overcome the drawbacks of these two approaches, recent works \cite{rastogi2019towards, zhang2019find} adopt a hybrid approach where some slots are {\it categorical} and others are {\it non-categorical}. 
For {\it categorical} slots the authors use a pick-list approach and employ span-prediction for {\it non-categorical} slots. 

Generation of slot values is typically based on Sequence to Sequence models that are also widely used in Machine Translation ~\cite{sutskever2014sequence,bahdanau2015neural}. This approach is ontology independent, flexible, and generalizable. 
~\citet{wu-etal-2019-transferable} proposes a sequence to sequence style model with soft copy mechanism ~\cite{see2017get}.
Our work is built upon TRADE \cite{wu-etal-2019-transferable}, due to its ontology free nature and flexibility to generate new values. \citet{kumar2020ma} also proposes a model based on TRADE with the addition of multiple attentions to model the interactions between slot and dialogue contexts. Team 2 at DSTC-8 \cite{LION-Net} also uses a TRADE-like model on Schema Guided Dialogues Dataset. This work can be considered as complementary to \cite{kumar2020ma} and \cite{LION-Net}, since our focus is on the weak-supervision setting. 

Recent research \cite{bingel2019domain} has also shown the efficacy of using weak dialogue level signals for domain transfer in DST. They pre-train a model with turn level signals on a domain and fine-tune with a dialogue-level reward signal on a target domain. In contrast, we propose to learn DST only from sparse signals available through API logs, without requiring any turn-level annotations.
\section{DST with Sparse Supervision}
In this section we formalize the task of DST with sparse supervision signals.
For a dialogue $D = \{ u_1, a_1, u_2, a_2, \cdots, u_{n_D}, a_{n_D}  \} $, we denote the agent and user utterances at turn $j$ as $a_j$ and $u_j$, respectively. 
In a multi-domain setup, let us assume we have  $S = \{ s_1, s_2, \cdots, s_N \}$ slots across all domains. For a dialogue turn $j$, the dialogue state ($y_j$) is defined as a set of slot-value pairs i.e. 
$y_j = \{(s_1, v_1), (s_2, v_2) \cdots (s_{N_j}, v_{N_j}) \}$,  where  $s_i \in S $ is a slot and $v_i \in V$ is its corresponding value.
Slots which have not been mentioned in the dialogue will have \textit{None} as their value in the dialogue state.
%are the slots and $V_j = \{v_1, v_2, \cdots, v_{N_j}\}$ are their corresponding values. 
%The number of slots in a dialogue state~($N_j$) could be the same as the total number of slots i.e. $N_j = N$ (in MultiWoz 2.1 dataset), or could be less ($N_j < N$) if the active domains for a dialogue are known and the values for only those slots need to be predicted (in SGD dataset).
Given the dialogue history of agent and user utterances up to turn $j$, 
the objective of a DST system is to predict the dialogue state $y_j$ at every turn $j \in n_D$ of the dialogue.

Existing DST systems consider this as a \emph{fully} supervised problem, where \emph{every} turn in the dialogue is manually annotated (set of labels $Y_D = \{y_1, y_2, \cdots, y_{n_D}\}$, $|Y_D|=n_D$). 
We focus on a realistic but harder setting, where supervision signals are available sparsely for only a few turns, i.e. %the number of labels for a dialogue is less than the number of turns i.e. 
$|Y_D|< n_D$. 
Such sparse labels for intermediate turns can be obtained when an API call is invoked after the turn (as illustrated in Figure \ref{fig:training-dst}). 
In industrial conversational systems, alignment of  API calls even at intermediate turns might not be available, 
as these are often logged at the whole dialogue level.
Such invocation logs can provide supervision only at the end of dialogue ( $Y_D = \{y_{n_D}\}$, $|Y_D| = 1$), pushing sparsity to the extreme.
%(in MultiWoz 2.1 dataset). 

It is important to note that training a DST model with sparse labels instead of turn-level annotations, makes this a weakly supervised task.
This is due to the fact that during training, the model has supervision signals for only a few turns ($y_j \in Y_D$), after which API calls were made. 
However, during inference it needs to predict the state at every turn in the dialogue $\{\hat{y_j}, \forall j \in (1, n_D)\}$.
In other words, this task requires building a model that can be trained with coarse supervision signal for multiple turns, but can make fine-grained predictions for individual turns. 

\section{Model Architecture}

\begin{figure}[t]
	\centering
	\includegraphics[width=\linewidth]{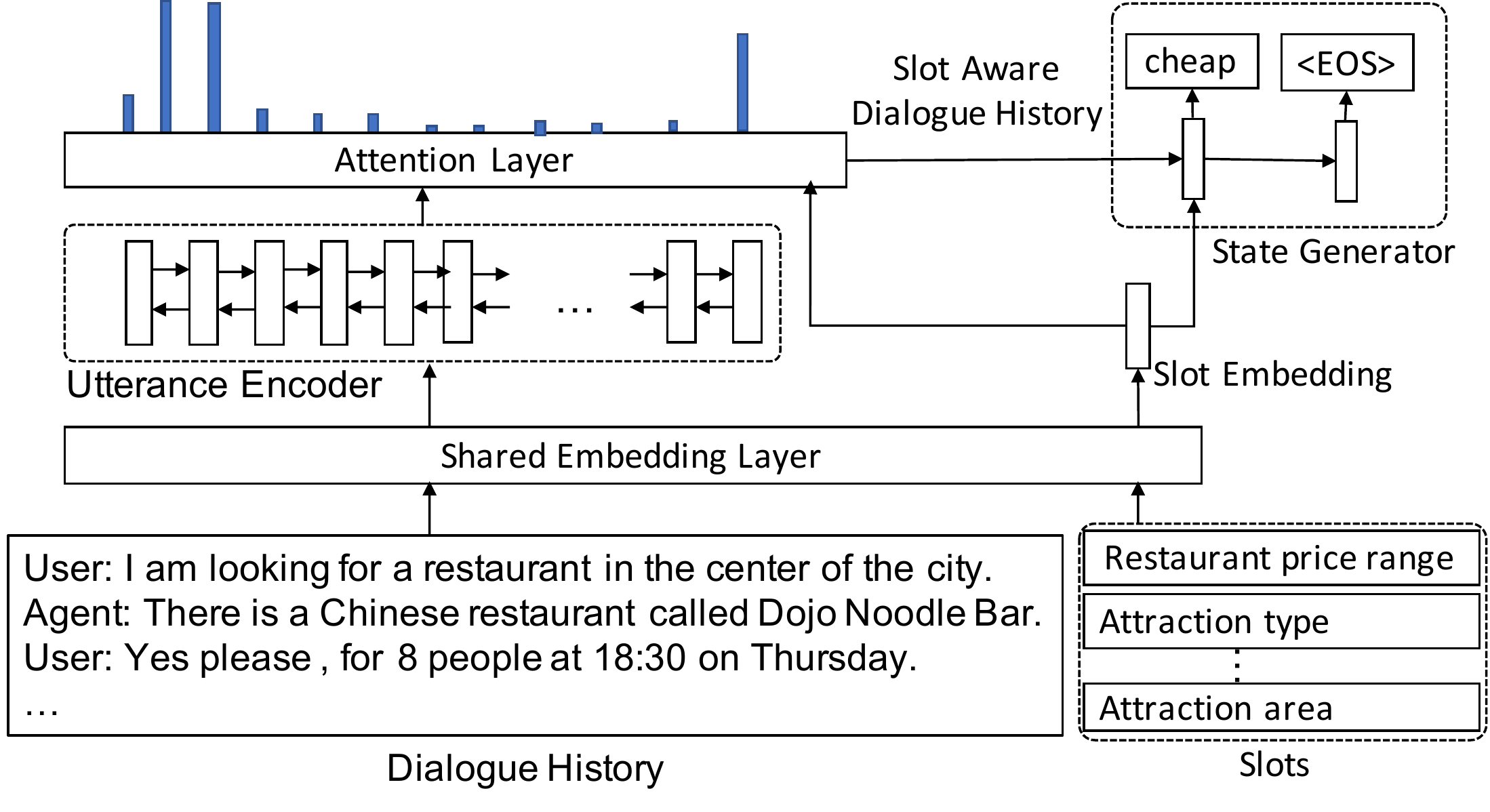}
	\caption{Model Architecture}
	\label{fig:model_arc}
	\vspace{-0.2cm}
\end{figure}

\begin{figure*}[t]
	\centering
	\begin{subfigure}{0.44\linewidth}
		\centering\includegraphics[width=\linewidth]{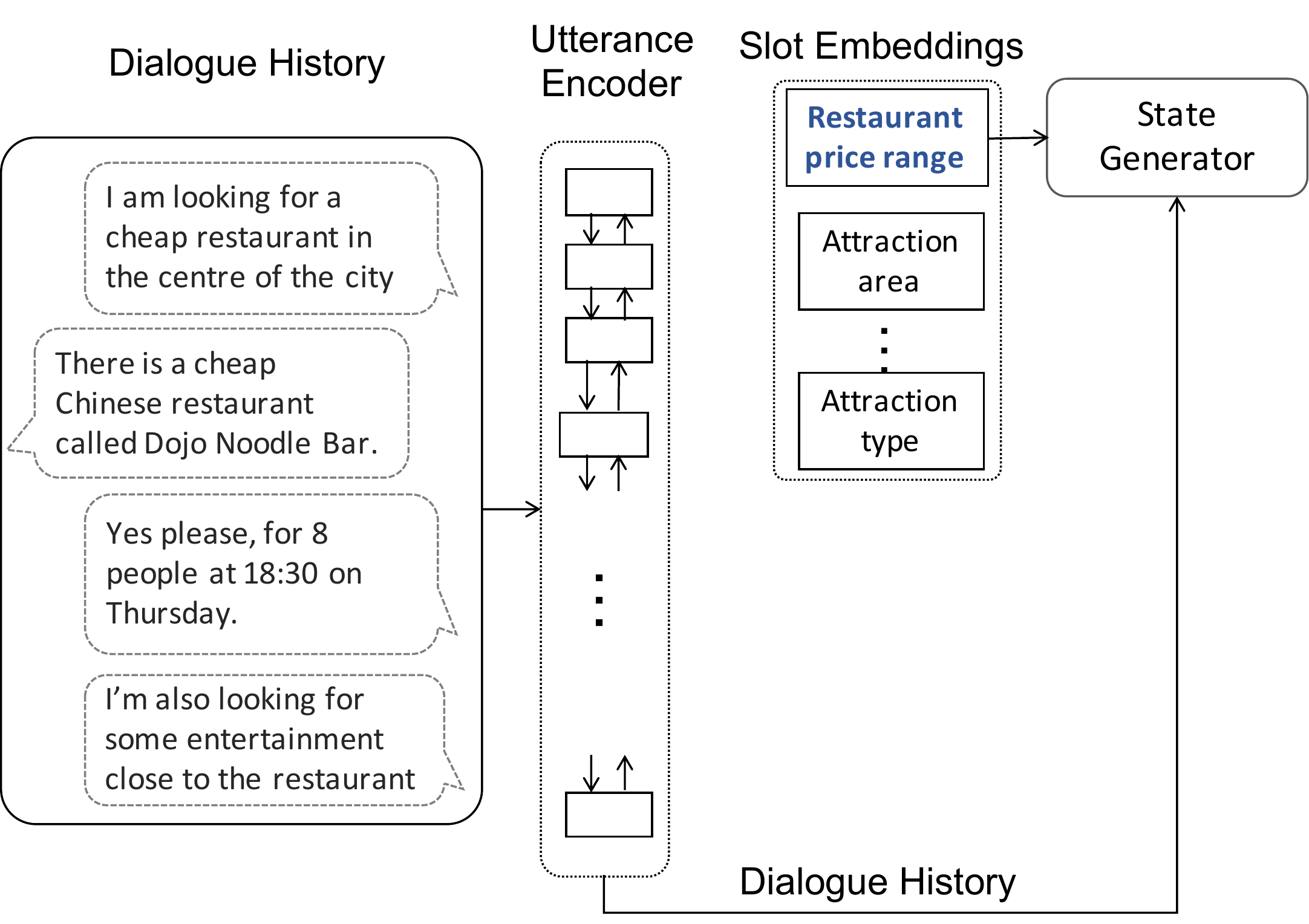}
		\caption{State Generator in TRADE}
		\label{fig:model_baseline}
	\end{subfigure}
	\hfill
	\begin{subfigure}{0.52\linewidth}
		\centering\includegraphics[width=\linewidth]{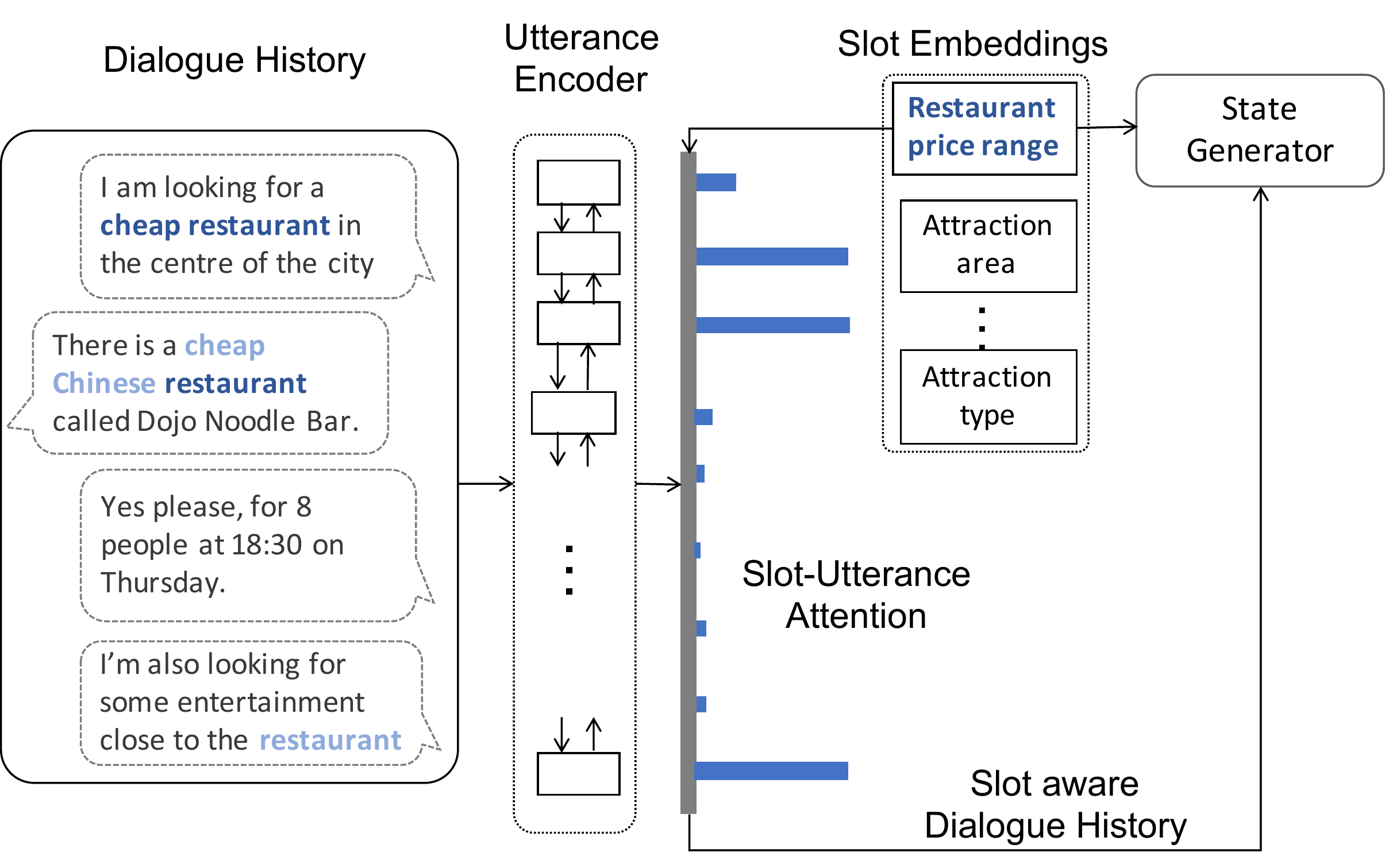}
		\caption{State Generator in AGG}
		\label{fig:model_aig}
	\end{subfigure}
\vspace{-0.1cm}
	\caption{Illustration of slot-utterance attention in AGG for state generator.  Instead of using the last RNN hidden state, a slot aware representation of dialogue history is used to initialize the state generator.}
	\label{fig:state_generator}
		\vspace{-0.2cm}
\end{figure*}

In this section we propose a model architecture for the dialogue state tracking task, as shown in Figure~\ref{fig:model_arc}. 
The architecture is based on the encoder-decoder paradigm and is adapted from TRADE~\cite{wu-etal-2019-transferable}. 
%and is based on the encoder-decoder paradigm inspired from TRADE~\cite{wu-etal-2019-transferable}.
Note that the model itself is agnostic to the learning setup (weak vs full supervision) used for the task.

%The model architecture is shown in Figure~\ref{fig:model_arc}.

\subsection{Utterance Encoder} 

Dialogue history at turn $j$ is formed by concatenating all agent and user utterances in the dialogue from turn $1$ to $j$. 
\begin{equation}
 X_j = {\mathrm{concat}} (u_1, a_1, u_2 , a_2, \cdots, a_{j-1} , u_j)
\end{equation}

where $\mathrm{concat}$  represents text concatenation function. The concatenated text is then tokenized using a white space tokenizer and embedded.

We use a combination of word and character embeddings to project all tokens to a low-dimensional vector space. We concatenate 300 dimensional  GloVe word embeddings~\cite{pennington2014glove} and 100 dimensional character embeddings \cite{hashimoto-etal-2017-joint} to represent the tokens. These embeddings are further fine-tuned during the training process.

The  dialogue history is embedded as:
\begin{equation}
E_j = \mathrm{embed}(X_j)  \in \mathbf{R}^{T_j \times d_{emb}} 
\end{equation}

where $T_j$ is the number of tokens in the dialogue history and $d_{emb}$ is the embedding dimension. 

For embedding domain and slot names, a separate randomly initialized embedding table is used in TRADE. 
In contrast, we share the embeddings among dialogue history, domain, and slot tokens by using the same $embed$ function for all of them. This ensures common tokens such as \textit{price range} have the same representation when they occur in a dialogue turn and as a slot name.

A bidirectional GRU \cite{cho-etal-2014-learning} is used to get a contextualized representation of dialogue history: 
\begin{align}
h^{enc}_t &=  \mathrm{BiGRU}(E_j[t]) \\
H^{enc}_j &= \{h^{enc}_1, h^{enc}_2, \cdots, h^{enc}_{T_j}\} \in \mathbf{R}^{T_j \times d_{hdd}} 
\end{align}

where $E_j[t]$ is the embedding for the $t^{th}$ token, $h^{enc}_t$ is the hidden state of the encoder at $t^{th}$ time-step, $d_{hdd}$ is the summation of output dimensions from forward and backward GRUs, and $H^{enc}_j $ represents the encoded dialogue history after $j$ turns.

\subsection{State Generator}

Another GRU is used as a decoder to generate slot values in the state generator.
For a dialogue turn $j$,  the decoder runs $N_j$ times to predict values for all $N_j$ slots in the dialogue state.

For a slot $s_k$ in dialogue state, the average  embeddings of its domain and slot name are used as input in the first time step of the decoder:

\begin{equation}
x^{dec}_0 = \mathrm{embed}(s_k)
\end{equation}

The output from the decoder at each time step is used for generating a word as the value of the corresponding slot. The generator may use words present in vocabulary or may choose to directly copy tokens from the source side, similar to the soft-gated copy mechanism \cite{see2017get}. This enables the model to predict slot values that might not have been pre-defined in the vocabulary but are present in the dialogue history.

\subsection{Attention Guided Generator (AGG)}

Unlike other sequence-to-sequence problems such as machine translation or summarization,
in dialogue state tracking the values to be generated are usually quite short - only one or two words long on average. 
Therefore, the initial hidden state of the generator is of crucial importance to generate accurate slot values. 

TRADE uses the last hidden state of the BiGRU in utterance encoder as initial hidden state of the decoder in state generator (shown in Figure \ref{fig:model_baseline}) .
\begin{equation}
 h^{dec}_0 =  h^{enc}_{T_j}
\end{equation}

$h^{enc}_{T_j}$ contains the whole encoded dialogue history up to turn $j$ and is agnostic of slots.

A dialogue history contains many turns and not all of those are important for a particular slot. In the weakly supervised setting, this becomes even more pronounced. 
In the hardest scenario, when supervision is available only for the last turn, the whole dialogue would be concatenated to create the dialogue history. 
This can result in a long sequence of text, where only a fraction of the information is relevant for a particular target slot.
%This makes the input text quite long with a large fraction of the available information being irrelevant for a particular target slot.
 
The model must learn to identify relevant turns for a target slot during training. This is because, during real-time deployment of the model at the inference stage, the model would need to predict the slot values at \textit{turn level}, although it has been trained with signals at \textit{dialogue level}. 
Therefore, a model relying on the complete dialogue history will struggle to make predictions during inference when only part of the dialogue is given.

We propose Attention-Guided-Generator (AGG) (shown in Figure~\ref{fig:model_aig}), which is initialized with a slot aware dialogue history. 
We apply a cross-attention between the target slot embedding and encoded source tokens to learn the relevance of tokens in dialogue history for that particular slot.
\begin{align}
\alpha_{s_k, t} &= \mathrm{attention}(s_k, h_t^{enc}) \label{eq:attn} \\ 
h_{s_k, 0}^{dec} &= \sum_{t=1}^{T_j} \alpha_{s_k, t} \cdot h_t^{enc}
\end{align}
where $\alpha_{s_k, t}$ denotes the attention weight for token $t$ in the dialogue for slot $s_k$, $h_{s_k, 0}^{dec}$ is the initial state of the decoder for the slot $s_k$. 

The use of this slot-aware dialogue history provides a better initialization for the generator.
More importantly, it also forces the model to focus only on those tokens in the dialogue history that are relevant to the slot to generate its value. This makes the prediction of values from partial input, i.e.~at the turn level, more robust and accurate.
 
\section{Experiments}
\label{exp}

\begin{table}[t]
	\centering
	\small
	\begin{tabular}{l c c}
		\hline
		& {\bf MultiWOZ 2.1}  & {\bf SGD} \\
		\hline\hline
		\# of domains & 7 & 16 \\
		\# of dialogues & 8,438 & 16,142 \\
		Total \# of turns & 113,556 & 329,964 \\
		Avg turns per dialogue & 13.46 & 20.44 \\
		\# of slots & 24 & 214 \\
		\hline
	\end{tabular}
	\caption{Dataset statistics for MultiWOZ 2.1 and Schema Guided Dialogues Dataset (SGD)}
	\label{table:dataset_statistics}
	\vspace{-0.3cm}
\end{table}

\begin{table*}[t]
	\centering
	\small
	\begin{tabular}{l| l | c c| c c }
		\hline
		\multirow{2}{*}{\bf Supervision} & \multirow{2}{*}{\bf Model } & \multicolumn{2}{c|}{\bf dev}  & \multicolumn{2}{c}{\bf test} \\
		& & {\bf SA } & {\bf Joint GA} & {\bf SA } & {\bf Joint GA } \\
		\hline\hline
		\multirow{3}{*}{Full} 
		& TRADE ~\cite{wu-etal-2019-transferable} & - & - & - & 45.60 \\
		& TRADE & 96.83 & 50.20 & 96.88 & 45.02 \\
		& {\bf AGG} & {\bf 97.22} & {\bf 52.40} & {\bf 97.25} & {\bf 47.57} \\
		\hline
		\multirow{2}{*} {Weak} &TRADE & 95.17 & 42.23 & 95.52 & 37.52 \\
		& {\bf AGG} & {\bf 96.10} & {\bf 48.03} & {\bf 96.30} & {\bf 43.46} \\            
		\hline
	\end{tabular}
	\caption{Results on MultiWOZ 2.1 for models trained with full and weak supervison. TRADE ~\cite{wu-etal-2019-transferable} refers to results reported in their paper.}
	\label{table:results_multiwoz}
	\vspace{-0.2cm}
\end{table*}

\begin{table}[t]
	\centering
	\small
	\begin{tabular}{l |c c| c}
		\hline
		\multirow{2}{*}{\bf Domain} & \multicolumn{2}{c|}{\bf Weak Supervision}  & \multirow{2}{*}{\bf Zero-Shot} \\
		& {\bf AGG } & {\bf TRADE } & \\ 
		\hline \hline
		taxi & 42.06 & 27.41 & 59.21 \\ 
		restaurant & 54.30 & 50.21 & 12.59 \\ 
		hotel & 37.56 & 31.84 &14.20 \\ 
		attraction & 57.25 & 46.53 & 20.06 \\ 
		train & 65.99 & 63.92 & 22.39 \\ \hline
	\end{tabular}
	\caption{Joint GA for Weak Supervision (AGG and TRADE) and Zero-Shot Transfer Learning (TRADE) on MultiWOZ 2.1 \cite{kumar2020ma}.}
	\label{table:zero_shot_multiwoz}
	\vspace{-0.3cm}
\end{table}

\subsection{Dataset}

We experiment on two datasets: MultiWOZ 2.1 \cite{eric2019multiwoz} and Schema Guided Dialogues (SGD) \cite{rastogi2019towards}. 
MultiWOZ  contains human-human written conversations spanning 7 domains. 
MultiWOZ 2.1 is the successor of MultiWOZ 2.0 \cite{budzianowski-etal-2018-multiwoz}, which fixes noisy state annotations.%, resulting in changes to over 32\% of annotations across 40\% of the dialogue turns. 
SGD is a larger dataset created by defining schemas and generating dialogue structures automatically. Humans were then asked to rewrite the dialogue structures in a more fluent natural language. 
For MultiWOZ, the predictions are made over all the domain-slot pairs, while in SGD predictions are made over a dynamic set of slots. 
The detailed statistics are shown in Table ~\ref{table:dataset_statistics}.

Currently, there is no publicly available large dataset with API call logs, therefore we use sparse dialogue state annotations for the weak-supervision setting.
In MultiWOZ dataset, only the dialogue states at the end of each dialogue are used for training to represent the hardest scenario for weak supervision, where API calls are fully characterized only at the end. 
%when all slots and values have been filled. 
SGD dataset has been annotated in a slightly different manner. If a particular service has been completed at a turn in the middle of the dialogue, say turn $j$, then its dialogue state annotations will be removed starting from turn $j+1$. Therefore, for each service, we only use the dialogue history up until the turn where it was completed.
This is analogous to an agent issuing API call to finish one task and then moving to the next task in a more complex dialogue.
Note that although the model is trained only with states of the last turn (for MultiWOZ) or few intermediate turns (for SGD), the dialogue states for all turns are still predicted and evaluated against the annotations.

\subsection{Implementation details}

We reimplement TRADE with PyTorch in Allennlp \cite{gardner-2017-allennnlp} framework based on the open-source code \footnote{https://github.com/jasonwu0731/trade-dst} and obtain results similar to the ones reported in the original paper.
On MultiWOZ 2.1 dataset, we follow the settings of TRADE as in the MultiWOZ 2.0.
%, which uses dialogues from 5 domains (restaurant, hotel, attraction, taxi, and train) and contains some tokenization and slot value fixes and canonicalization. In the preprocessing stage, the tokens are all lower-cased. 
On SGD dataset, we tokenize the dialogue history with Spacy tokenizer and preserve casing. 
For the encoder, one layer bidirectional GRU with 400 hidden units is used, and a dropout of 0.2 is applied to the embedding output as well as the GRU output. For the decoder, we use teacher forcing \cite{NIPS2015_5956} ratio of 0.5 to sample from gold and generated tokens, and greedy decoding is used due to the short length of the target sequences. Adam optimizer~\cite{kingma2014adam} is used with an initial learning rate of 0.001.
% and a batch size of 32. 
%We stop model training after the joint goal accuracy on the dev set has not improved for 6 epochs. 
All reported numbers are produced by running with 5 random seeds and averaging their results.

\begin{figure*}[t]
	\centering
	\includegraphics[width=0.8\linewidth]{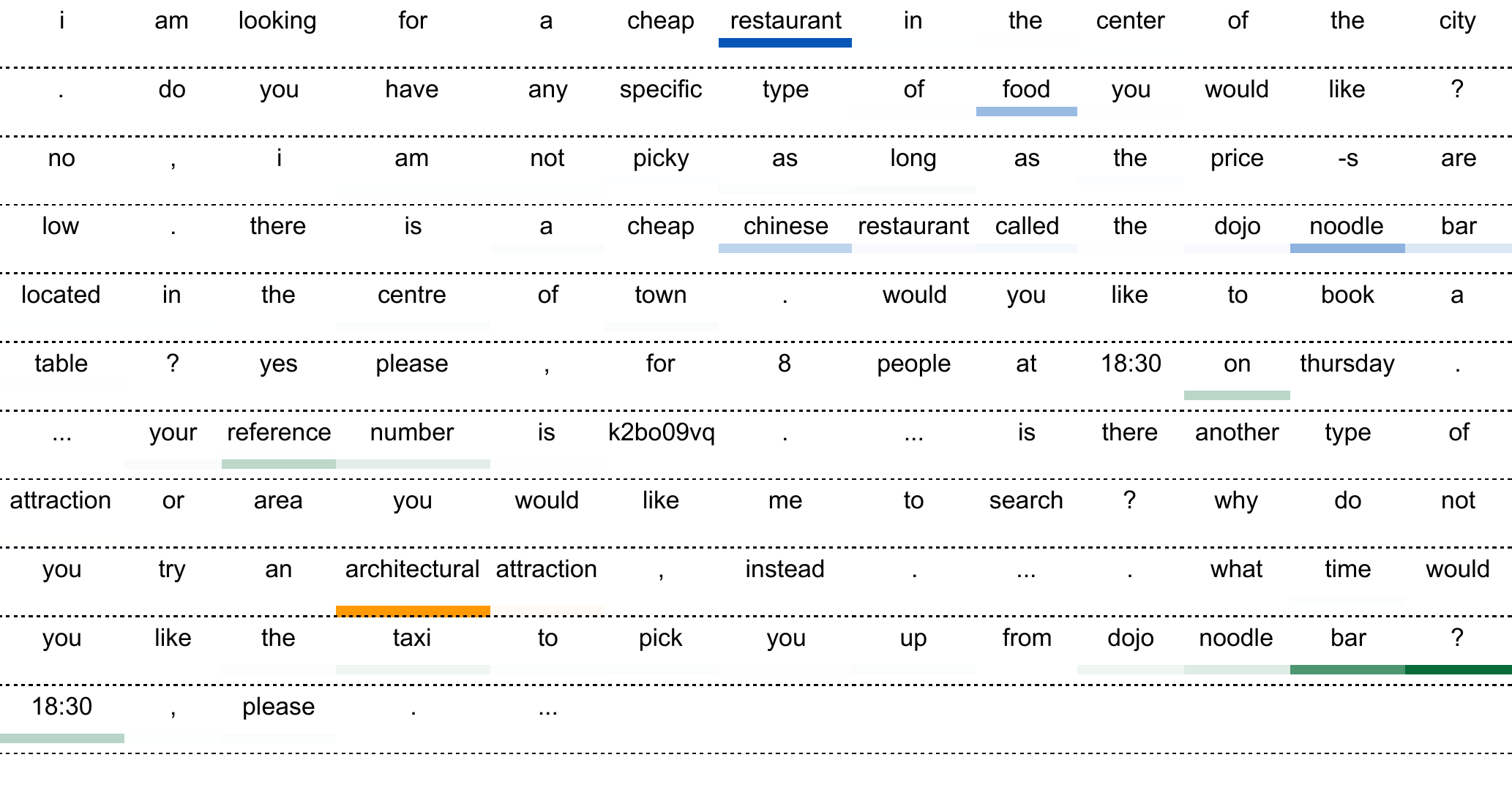}
	\vspace{-0.2cm}
	\caption{Attention weights (calculated from Equation~\ref{eq:attn}). for a sample dialogue from MultiWOZ 2.1. Color bars under the texts represent attention weights for slot \textcolor{slot-blue}{\it restaurant-price\_range}, \textcolor{slot-green}{\it taxi-leave\_at} and \textcolor{slot-orange}{\it attraction-type}.}
	\label{fig:slot_attention}
\end{figure*}

\begin{table*}[ht]
	\centering
	\small
	\begin{tabular}{l  l |  c c | c c | c c}
		\hline
		\multirow{2}{*}{\begin{tabular}[c]{@{}l@{}}{\bf Supervision} \\ {\bf Type} \end{tabular}} & \multirow{2}{*}{\bf Model} & \multicolumn{2}{c|} {\bf All Services}& \multicolumn{2}{c|}{\bf Seen Services} & \multicolumn{2}{c}{\bf Unseen Services}\\
		& & {\bf Joint GA } & {\bf Average GA} & {\bf Joint GA } & {\bf Average GA} & {\bf Joint GA } & {\bf Average GA}  \\ 
		\hline\hline
		\multirow{4}{*}{Full} & Google Baseline & 0.2537 & 0.5605  & 0.4125 & 0.6778 & 0.2000 & 0.5192 \\
		& Team 2  & 0.3647 & 0.7438 & 0.7363 & 0.9132 & 0.2406 & 0.6850 \\
		& TRADE  & 0.3547 & 0.6429 & 0.7133 & 0.8885 & 0.2350 & 0.5576 \\
		& AGG & 0.3273 & 0.6192 & 0.7196 & 0.8902 & 0.1964 & 0.5250 \\
		\hline
		\multirow{2}{*}{Weak} &	TRADE & 0.2139 & 0.5766 & 0.4791 & 0.7968 & 0.1253 & 0.5000 \\
		&	{\bf AGG} & {\bf 0.2501} & {\bf 0.6261} & {\bf 0.5648} & {\bf 0.8522} & {\bf 0.1452} & {\bf 0.5474} \\
		\hline
	\end{tabular}
	\caption{Experiment results on SGD. Google Baseline and Team 2 results are from ~\cite{rastogi2020schema} and ~\cite{LION-Net}. {\it fs} and {\it ws} refer to models trained with full and weak supervision respectively.}
	\label{table:results_sgd}
	\vspace{-0.2cm}
\end{table*}

\subsection{Results on MultiWOZ 2.1}
\label{multiwoz-results}
There are two evaluation metrics for MultiWOZ 2.1 dataset.

\noindent \textbf{Slot accuracy (SA): } compares the predicted value for each slot at each dialogue turn. 

\noindent \textbf{Joint Goal Accuracy (Joint GA): } computes accuracy of the predicted dialogue state at each dialogue turn. The predicted state is considered to be correct only if all slot values are correctly predicted.

Experiment results on the test set of MultiWOZ 2.1  are shown in Table~\ref{table:results_multiwoz}. 
Values for Slot Accuracy are quite high for all methods, since many slots have {\it none} values across the dialogue turns, therefore we focus more on the {\it Joint goal accuracy} metric. 
We first observe  that AGG outperforms TRADE in both full and weak supervision settings comfortably, with improvements of 1.97\% and 5.94\% absolute, respectively. 

We also note that  AGG trained with weak supervision is only 4.11\% lower in performance than the model trained with full supervision. 
This performance drop is much lower compared to TRADE when used where the gap in performance between full and weak supervision is  8.08\%. This shows the effectiveness of slot aware attention in sparse settings.
The low drop in performance due to weak supervision illustrates that using broadly available, inexpensive labels we can achieve performance very close to a model trained with full supervision.

\noindent \textbf{Scaling to new domains: }
In the next set of experiments, we evaluate the efficacy of using the weak supervision approach for scaling to new domains. 
For a new domain, we propose to use automatically-collected sparse signals for weak supervision, instead of predicting with zero-shot learning. 
We compare our weakly supervised model (AGG-ws) on MultiWOZ 2.1, with a standard approach from  literature, i.e. TRADE trained on turn-level signal evaluated on a new domain in zero-shot setting. 

We measure Joint GA on test sets only from a single target domain. 
Training set contains dialogues from the other four domains with full supervision for \textbf{Zero-Shot}, while for {\bf Weak-Supervision} setting the training set contains dialogues with weak supervision from all five domains.

%As the results in Table \ref{table:zero_shot_multiwoz} indicate, using weak supervision helps us achieve much higher accuracy for the target domains than a zero-shot setting for inference.

From the results in Table \ref{table:zero_shot_multiwoz}, we first note that  AGG consistently outperforms TRADE for all domains in the weakly supervised setting.
Furthermore, we show that using weak supervision signal for all domains is better than turn level annotations in source domains, followed by zero shot prediction in target domain. 
Due to the easy availability of weak signals in practical applications, weak supervision seems a more appealing choice than zero shot learning for scaling DST across domains.

\noindent \textbf{Slot-Utterance attention analysis: }
To validate the effectiveness of the slot-attention mechanism in the weak supervision setting, we run inference on a few randomly selected multi-domain dialogues from MultiWOZ 2.1 dataset. For a sample dialogue history, the attention weights of AGG for some slots are plotted in Figure~\ref{fig:slot_attention}. The weights show that the model can focus on the most relevant parts of the long dialogue history for different slots. 
For instance, for slot {\it restaurant-price\_range}, the model focuses on {\it restaurant}, {\it food}, {\it chinese restaurant} etc., which are relevant to the slot. 
This helps to better initialize the state generator and makes the generation task much easier to learn.

\begin{table*}[t]
	\centering
	\small
	\begin{tabular}{l l | c c | c c}
		\hline
		\multirow{2}{*}{\bf Supervision} &	\multirow{2}{*}{\bf Model} & \multicolumn{2}{c|}{\bf MultiWOZ 2.1}  & \multicolumn{2}{c}{\bf SGD}  \\
		& & {\bf Epoch Time } & {\bf Training Time } & {\bf Epoch Time } & {\bf Training Time} \\ 
		\hline\hline
		\multirow{2}{*}{Full} 
		& TRADE &	00:03:20 &	01:15:52 &	00:14:52 &	05:34:04 \\
		& AGG & 00:03:45 & 01:44:25 & 00:16:43 & 05:50:29 \\
		\hline
		\multirow{2}{*}{Weak} 
		& TRADE & 00:00:50 & 00:22:33 & 00:03:37 & 01:57:26 \\
		& AGG & 00:00:51 & 00:37:45 & 00:04:08 & 01:25:25 \\
		\hline
	\end{tabular}
	\caption{ Model epoch training time and total training durations for different supervision settings.}
	\label{table:training_time}
	\vspace{-0.3cm}
\end{table*}

\subsection{Results on SGD}

For evaluation on SGD dataset, we use the official evaluation metrics \cite{rastogi2019towards}.

\noindent \textbf{Joint Goal Accuracy (Joint GA): } Same as for MultiWOZ, but only the slot values for \emph{active} services need to be predicted correctly. %This is different from MultiWOZ where slots for \emph{all} services are considered.
%, i.e. slots for inactive services need to be correctly predicted as \emph{none}.

\noindent \textbf{Average Goal Accuracy (Average GA): }  Average accuracy of predicting active slot values correctly. Similar to Slot Accuracy in MultiWOZ, but slots with \emph{none} values are not counted.

%where ground truth values are \emph{none}, are not counted.

We compare AGG with our implementation of TRADE trained and evaluated on the SGD dataset on both full and weak supervision settings. We also include results from the Baseline system  and one of the participant teams (Team 2) in DSTC8 competition \cite{rastogi2020schema}.
We compare with Team 2 since they did not use large scale  pre-trained contextualized embeddings such as BERT \cite{devlin-etal-2019-bert} and their model architecture is the most similar to ours. 
Using pre-trained embeddings for better initialization is complementary to our approach and not the focus of this work, so we leave that exploration for future work.

From the results on SGD test set as shown in Table~\ref{table:results_sgd}, we first note that in \emph{weak supervision} setting, the proposed model AGG outperforms TRADE by large margins for both {\bf Seen} and {\bf Unseen Services}. 
Furthermore, the drop in performance for training with weak vs full supervision, is consistently lower for AGG than for TRADE. This is inline with our observation from results on MultiWOZ dataset as well (in Section \ref{multiwoz-results}).  
On {\bf Seen Services}, 
AGG trained with weak supervision outperforms the Google Baseline trained with full supervision. 

%Our implementation of TRADE achieves results similar to Team 2, which uses additional losses such as the loss for predicting requested slots and active intents. In the fully supervised setting, AGG-fs outperforms TRADE-fs by a slight margin. 
%However, under weak supervision, AGG-ws improves over TRADE-ws by a large margin of 8.57\% for Joint Goal Accuracy. 
We note that with \emph{full supervision} the performances of TRADE and AGG 
are lower than that of Team 2. 
One factor could be that Team 2 also uses the service and slot descriptions provided in the SGD dataset. This additional textual information possibly helps quite a bit in informing the model. However, we only use service and slot names in our models to make the learning task similar across datasets, i.e. same as MultiWOZ where such additional signals are not available.

%
%
%\begin{figure}[t]
%	\centering
%	\begin{subfigure}{0.8\linewidth}
%		\centering\includegraphics[width=\linewidth]{figures/dev_learning_curve_multiwoz.pdf}
%		\caption{MultiWOZ 2.1}
%	\end{subfigure}
%	\begin{subfigure}{0.8\linewidth}
%		\centering\includegraphics[width=\linewidth]{figures/dev_learning_curve_sgd.pdf}
%		\caption{SGD}
%	\end{subfigure}
%	\caption{Learning curves for {\it TRADE}, and {\it AGG} on validation sets. {\it -fs} and {\it -ws} denote the model versions trained with full and weak supervision, respectively.}
%	\label{fig:learning_curve}
%	\vspace{-0.3cm}
%\end{figure}

\subsection{Efficiency Analysis}

In this subsection, we analyze the efficiency of training weakly supervised models compared to traditional fully supervised ones from time and data requirements perspective.

\subsubsection*{Training Time}

The training time for different model variations is shown in Table~\ref{table:training_time}. All models have been trained on an  AWS EC2 p3.16xlarge instance which has 8 Tesla V100 GPUs.
We can see that using only weak supervision, the time per epoch and time required for complete model training has reduced to nearly a \emph{quarter} on both MultiWOZ 2.1 and SGD datasets. 

%
%\subsubsection*{Learning curve}
%
%
%We plot the Joint Goal Accuracy against the epoch number in Figure~\ref{fig:learning_curve} to show the training progression of different models. The learning curves prove that AGG has a jump-start over TRADE from the first epochs, which shows the strength of the slot-aware initialization of the decoder hidden state. Ultimately, AGG converges to a better state in both the full and weakly supervised setup on both datasets.

\subsubsection*{Data Efficiency}

\begin{figure}[t]
	\centering
	\includegraphics[width=0.95\columnwidth]{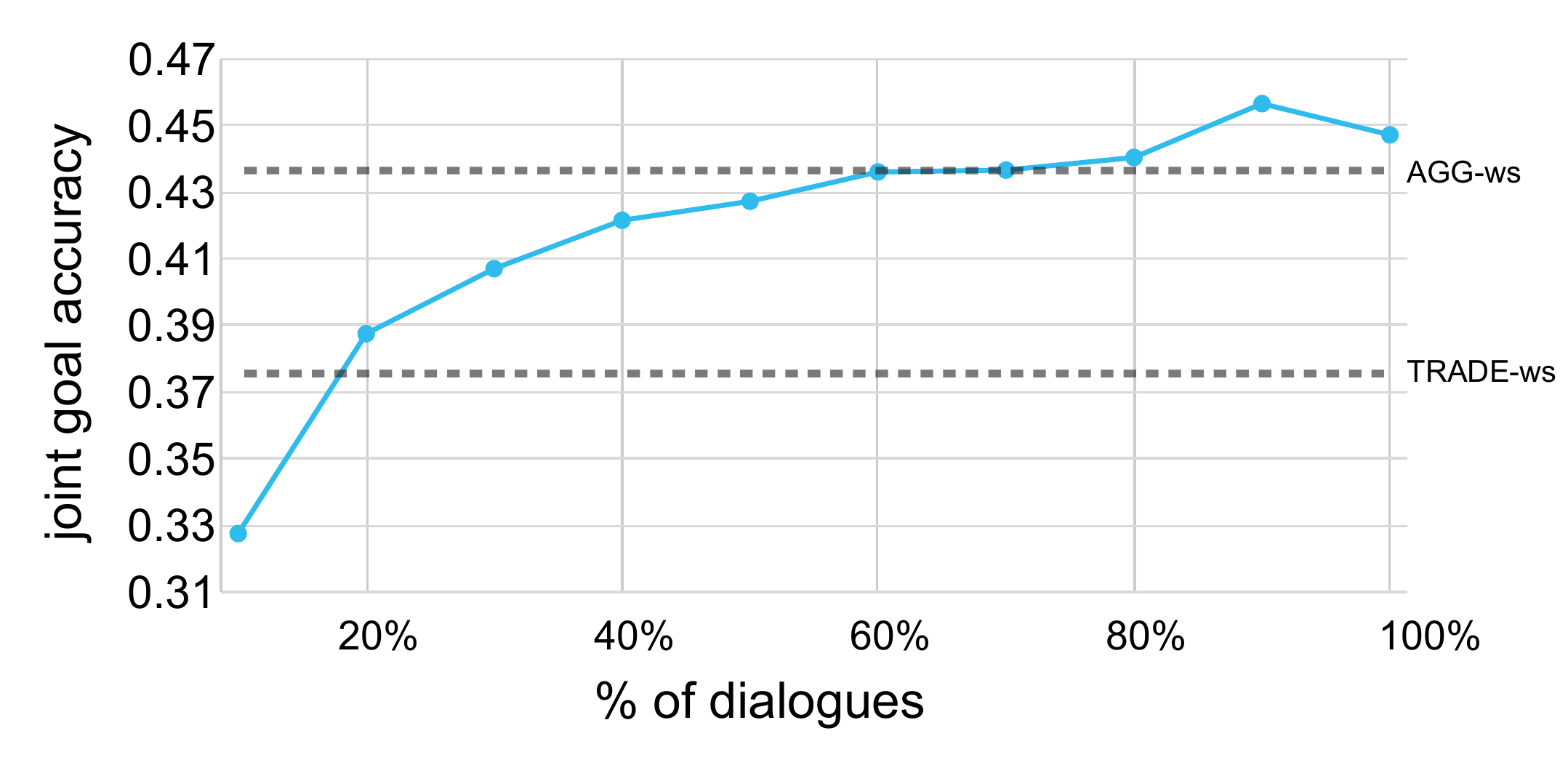}
	\caption{TRADE model results on test split of MultiWOZ 2.1 trained on subsampling dialogues at different rates. For reference, the upper and lower dashed lines are results of the {\it AGG} and {\it TRADE} trained on all dialogues but in weakly-supervised setting.}
	\label{fig:data_efficiency}
		\vspace{-0.3cm}
\end{figure}

We want to analyze the data efficiency of weak supervision and our model in terms of impact on performance  with varying data sizes. We subsample the dialogues in MultiWOZ 2.1 and plot the result of TRADE trained with full supervision on varied dataset sizes.  
As Figure~\ref{fig:data_efficiency} shows, ~20\% of dialogues are required with turn level supervision to match the performance of a TRADE model that is trained with weak signals on the full dataset. 
%On the other hand, AGG-ws learns a strong model and we need 60\%-80\% of the dialogues with full turn level supervision to reach the performance bar of the attention guided model. 
On the other hand, AGG learns a strong model from weak state annotations that is on par with TRADE trained on 80\% of dialogues annotated at every dialogue turn.
This demonstrates that AGG is more data-efficient and particularly effective for weakly supervised DST.

\section{Conclusion}

In this work, we introduce a new learning setup for dialogue state tracking by effectively leveraging sparse dialogue states as weak supervision signals.
We propose an encoder-decoder based approach inspired by a copy-generative model TRADE. 
We introduce an attention-guided generator that learns to focus on relevant parts of a dialogue for a given slot to make predictions.
Experimental results on two large open-domain dialogue datasets show that training on sparse state annotations can achieve a predictive performance close to that when training on full data. 
This work follows the body of literature aiming to train models with fewer labeled samples. 
Our approach shows that, without manual annotations and using only proxy signals for weak supervision, we can train models at large scale and expand to new domains.
\bibliography{liang}
\bibliographystyle{acl_natbib}

\end{document}